\documentclass[runningheads]{llncs}

\newif\ifreview
\reviewfalse           

\ifreview
  \usepackage[review,year=2026,ID=0]{eccv}
\else
  \usepackage{eccv}
\fi

\usepackage{eccvabbrv}

\usepackage{graphicx}   
\usepackage[expansion=false]{microtype}  
\usepackage{booktabs}   
\usepackage{float}      
\usepackage{tikz}       
\usetikzlibrary{arrows.meta, positioning, shapes.geometric, calc, fit, backgrounds}

\usepackage[accsupp]{axessibility}

\ifreview
  \usepackage[pagebackref,breaklinks,colorlinks,citecolor=eccvblue]{hyperref}
\else
  \usepackage{hyperref}
  \hypersetup{hidelinks}
\fi

\usepackage{orcidlink}


\usepackage{xcolor}

\makeatletter
\@ifundefined{maketitleold}{}{%
  \renewcommand{\maketitle}{%
    \author{Anonymous CDEL Workshop Submission -- ECCV \eccv@year{}}%
    \titlerunning{ECCV \eccv@year{} CDEL Workshop Submission}%
    \authorrunning{ECCV \eccv@year{} CDEL Workshop Submission}%
    \institute{}%
    \maketitleold}%
}
\makeatother

\begin{document}

\newcommand{\AcceptanceNote}{Accepted at the Curated Data for Efficient
  Learning (CDEL) Workshop,\\ held in conjunction with the European Conference
  on Computer Vision\\ (ECCV 2026), Malm\"o, Sweden.}

\newcommand{\FundingAcknowledgement}{This work was supported by the Belgian
  Defense under Grant DAP 23/08.}

\newcommand{\PaperTitle}{Diffuse the Object, Keep Its Label:\\
Curating Detector Training Data from\\
a Few Unlabeled Photographs\\
via VLM-Built 3D Vegetation Scenes}
\title{\PaperTitle}

\titlerunning{Diffuse the Object, Keep Its Label}

\newcommand{\CorrespondingAuthorMark}{\textsuperscript{\textdagger}}
\author{Mario Malizia\inst{1,2,3}\orcidlink{0009-0008-6618-5059}\CorrespondingAuthorMark \and
Marnix Enting\inst{1}\orcidlink{0009-0000-5360-4076} \and \\
Rob Haelterman\inst{1}\orcidlink{0000-0002-1610-2218} \and
Ken Hasselmann\inst{1}\orcidlink{0000-0002-8196-9889}}
\authorrunning{M.~Malizia et al.}
\institute{Royal Military Academy, Brussels, Belgium\\
\email{\{mario.malizia,marnix.enting,rob.haelterman,ken.hasselmann\}@mil.be}
\and KU Leuven, Leuven, Belgium
\and Flanders Make, Heverlee, Belgium}

\maketitle

\ifreview\else
\begingroup
\makeatletter
\renewcommand{\thefootnote}{}
\renewcommand{\@makefntext}[1]{\noindent\footnotesize #1}
\makeatother
\begin{NoHyper}
\footnotetext{%
  \AcceptanceNote\par
  \FundingAcknowledgement\par
  \CorrespondingAuthorMark\ Corresponding author.%
}
\end{NoHyper}
\endgroup
\fi

\begin{abstract}
  Labeled images of small objects hidden in vegetation are scarce, and
  detectors trained on them generalize poorly across sites.
  Rather than reusing labels collected at another site, we synthesize
  labeled training images from a handful of unlabeled photographs of the
  deployment site itself.
  A vision--language model generates a coarse 3D vegetation scene from one
  photograph; placing 3D object meshes in the scene yields bounding boxes,
  segmentation masks, and per-instance occlusion directly from the scene
  geometry, without manual annotation.
  A lightweight adapter fine-tuned on the photographs conditions a
  diffusion pass that re-textures the renders, and a graded mask-lock sets
  how much diffusion may touch the object itself.
  In our runs this grade was the most influential curation choice:
  lightly diffusing the object improves minority-class recall over fully
  protecting its pixels, while unrestricted diffusion dissolves it.
  Trained on these images, a standard detector matched or exceeded its
  counterpart trained on a larger labeled dataset of real images from a
  different site, consistently across seeds on a humanitarian-demining benchmark;
  the comparison is thus unsupervised site adaptation from a handful of
  photographs against conventional cross-site label reuse.
  In our ablations the gains were largely insensitive to the photograph
  and crop budgets, and in-domain accuracy did not predict cross-site
  performance.
\end{abstract}

\section{Introduction}
\label{sec:intro}

Detecting small or camouflaged objects in natural vegetation is a
recurring problem across safety-critical and resource-constrained domains,
from humanitarian landmine
clearance~\cite{MineInsight2025,Vivoli2024SULAND,Hasselmann2024MultiRobot}
to invasive species monitoring~\cite{Doherty2024LeafySpurge}, precision
agriculture~\cite{Steininger2023CropAndWeed}, and wilderness search and
rescue~\cite{Kim2026ForestPersons}.
These tasks share a bottleneck: annotated data are scarce, expensive
to collect, and often physically hazardous to acquire.

Synthetic images are a natural response to this scarcity, but
existing approaches have well-documented limits.
Procedural 3D rendering in open-source engines such as
Blender~\cite{Blender}, used for detector training via domain
randomization~\cite{Tremblay2018DR,Prakash2019SDR}, offers full
control over object pose, occlusion, and label geometry, yet without
extensive artist input its output looks synthetic: flat foliage
textures, uniform shading, and no sensor noise.
Neural reconstructions based on 3D Gaussian
Splatting~\cite{Michiels2025CutSplat,Zanjani2025GSDataGen}
produce visually convincing backgrounds but require dense captures of the
target environment, which are often unavailable in the domains listed
above.
Text-to-image diffusion models~\cite{Rombach2022LDM} yield photorealistic
outputs from prompts but do not natively expose the pose and label
information needed for detector training, nor can they exploit an
available 3D model of the object.

A promising middle ground is to render a coarse 3D scene and then refine
its appearance with a diffusion model.
Recent instances of this pattern pair Blender rendering with a
depth-conditioned diffusion model for mushroom
segmentation~\cite{Karoly2025Mushroom}, refine Skinned Multi-Person
Linear (SMPL) human renders with an edge-conditioned Stable Diffusion
model for privacy-compliant human
synthesis~\cite{Patwari2024RefSD}, and insert 3D vehicle assets into
Gaussian-splatting scenes via a one-step diffusion model for driving
simulation~\cite{Ljungbergh2025R3D2}.
IntrinsicControlNet~\cite{Lu2025Intrinsic} conditions generation on
albedo, normal, and lighting buffers extracted from a renderer, and
Coarse-to-Real~\cite{GomezNogales2026C2R} extends the idea to
game-engine crowd videos.
In parallel, LLM- and VLM-driven agents synthesize executable Blender
code as a scene scaffold that a diffusion model can then refine:
SceneCraft~\cite{Hu2024SceneCraft} translates a text description into
Blender code through scene-graph planning, and
SEIG~\cite{He2026SEIG} goes further by reconstructing a Blender scene
from a single reference photograph via staged inverse graphics.

We build on both lines and address a deployment setting common in
humanitarian mine action: the operator knows, from a non-technical
survey~\cite{IMAS0810} and historical clearance
records~\cite{Saliba2024MineTypeGIS}, both the terrain type of the
suspected hazardous area and the family of ordnance likely to be
encountered.
This prior knowledge, standard practice in mine action doctrine,
makes a site-specific synthetic dataset feasible without
large-scale per-site data collection: the terrain distribution can be
captured by a few unlabeled reference photographs, and the object
catalog reduces to a handful of known mesh models.
The conventional alternative is to reuse labeled data collected at
previously surveyed sites, and it is exactly this cross-site reuse
that breaks down: every new suspected hazardous area is, in effect,
a new visual domain for which labels do not exist.
Generating the training set from a few photographs of the site
itself sidesteps the reuse problem and can be repeated for each
new site at negligible data-collection cost.

Our pipeline proceeds as follows.
A VLM reads one reference photograph and emits a coarse Blender
scene of the terrain; the scene is populated with 3D meshes of the
expected ordnance (in our study, four PFM-1 butterfly landmine
variants and one PMA-2 starfish anti-personnel landmine), so every
render carries bounding boxes, masks, and per-instance occlusion by
construction.
A Low-Rank Adaptation (LoRA)~\cite{Hu2022LoRA}, fine-tuned on
unlabeled crops of the target terrain, then guides
SDEdit~\cite{Meng2022SDEdit} to replace surface appearance while
preserving the 3D layout, and a sensor-noise pass matches the target
camera response.

Existing render-then-refine pipelines assume that the diffusion model
can plausibly re-texture the object; this holds for mushrooms,
vehicles, and human bodies, but becomes fragile for objects that are
rare or absent from generative training corpora, such as
ordnance: unconstrained diffusion can distort, remove, or hallucinate
around the object and invalidate the associated label.
We enforce preservation directly through \emph{mask-locked
compositing}: the diffusion output is used only for the background,
while the object pixels rendered by Blender are pasted back through
the render's visible-surface (post-occlusion) mask, so occluding
vegetation stays in front of the object after the paste.
The label therefore remains valid by construction.

A further property separates this setting from prior render-and-refine
applications: surface-laid objects often lie partially \emph{embedded
in} vegetation rather than solely on top of it.
Occlusion by vegetation is not a marginal effect but a primary driver of
detection failure in minefield environments~\cite{Baur2024VegCoverage},
and a generative model alone cannot reproduce it with the
supervision detection requires: diffusion exposes neither the
visible/amodal mask split nor the coverage ratio, and thin occluders
such as grass blades fall below the effective resolution of
latent-space conditioning.
In our pipeline, occlusion is instead inherited from the 3D scene:
the vegetation primitives generated by the VLM geometrically cover
the object, the visible-surface mask is exact, and occlusion is
controlled per instance.

We instantiate the pipeline on surface-laid landmine detection in
vegetated terrain, a data-scarce and safety-critical
detection task in computer vision.
Existing benchmarks~\cite{MineInsight2025,Vivoli2024SULAND} together
provide only a few tens of thousands of annotated images across a
handful of terrain types, and expanding them requires inert ordnance
and controlled field sites.
The pipeline is not tied to this task in principle: other detection
settings in which the object catalog is small, geometrically well
defined, and hard to acquire in real data could substitute their own
3D mesh library and terrain LoRA, though we do not evaluate that
generalization here.

\noindent\textbf{Contributions.}
\begin{enumerate}
  \item A render-and-refine pipeline for curating detector training
        data in vegetated terrain: a VLM generates a coarse Blender scene
        from one reference photograph, and a LoRA trained on unlabeled
        crops of the same site adapts the diffusion prior.
        A single graded mask-lock parameter~$\alpha$ acts as a curation
        dial: it sets how much diffusion may touch the object, while
        mask-locked compositing preserves the placed 3D meshes and
        their labels.
  \item A preliminary cross-site generalization study on surface-laid
        landmine detection against the same detector trained on a far
        larger labeled dataset of real images: in our runs, varying the mask-lock
        grade alone shifted generalization more than any data-budget
        choice (Sec.~\ref{sec:experiments}).
\end{enumerate}
\section{Related Work}
\label{sec:related}

\paragraph{Synthetic data for object detection.}
Generating training images from 3D engines is a well-established strategy.
Domain randomization varies lighting, texture, and geometry during
rendering to broaden the effective support of the training
distribution~\cite{Tobin2017DR,Tremblay2018DR}.
Structured variants impose scene-level priors such as object placement
rules and context-aware composition~\cite{Prakash2019SDR}.
A complementary line replaces the renderer with a 2D generative model
and relies on compositing for the background.
In agriculture, Generate-Paste-Blend-Detect~\cite{Giakoumoglou2023GPBD}
generates whitefly instances with a denoising diffusion probabilistic
model (DDPM) and composites them onto real backgrounds using Poisson
or Gaussian blending to train YOLOv8.
Modak~\etal~\cite{Modak2025SDWeed} apply Stable Diffusion inpainting to
augment weed images under model quantization constraints, showing that
generative augmentation remains effective on edge hardware.
Patch-level synthesis~\cite{Li2025PatchSynth} composites species-level
crop and weed instances onto real backgrounds for fine-grained
segmentation.
These 2D approaches produce plausible images but lack geometric grounding:
object scale, occlusion, and viewpoint are controlled only implicitly, and
annotations must be derived from the generation mask rather than from
known 3D geometry.

\paragraph{3D rendering with diffusion refinement.}
A growing line of work addresses the appearance gap between renders
and real images by rendering a
coarse 3D scene and then transforming it into a photorealistic image
with a diffusion model.
The closest system to ours is that of K\'{a}roly and
Galambos~\cite{Karoly2025Mushroom}, who regenerate entire
Blender-rendered mushroom scenes with depth-conditioned Stable
Diffusion XL (SDXL), reaching F1\,=\,0.859 on the M18K benchmark
with only synthetic training data.
Full regeneration is viable in their setting because the target class
is well represented in generative corpora (their object LoRA is
trained largely on web images), yet they report that the depth
conditioning occasionally admits unlabeled instances that bias the
trained model toward precision over recall.
Notably, on their own real-world data the diffusion-refined training
set does not outperform the plain Blender renders, leaving open
whether generative refinement helps precisely where it should matter
most: under domain shift.
RefSD~\cite{Patwari2024RefSD} pseudonymizes humans by rendering an
SMPL mesh over each subject and passing it through SDXL with a Canny
ControlNet, so the 3D scaffold preserves pose while the diffusion
redraws surface appearance under text-prompt attribute control.
R3D2~\cite{Ljungbergh2025R3D2} takes a different route: it inserts 3D
vehicle assets into Gaussian-splatting reconstructions of driving scenes
and uses a one-step diffusion model to add shadows and lighting
integration rather than to re-texture the scene.
Coarse-to-Real~\cite{GomezNogales2026C2R} extends the render-then-refine
idea to video, learning a neural renderer that transforms game-engine
crowd simulations into photorealistic sequences.
IntrinsicControlNet~\cite{Lu2025Intrinsic} conditions generation on
intrinsic buffers (albedo, normals, lighting) extracted from a rendering
engine, bridging synthetic and real distributions in a single model.

A common assumption across these systems is that the diffusion model can
plausibly re-texture the target category: mushrooms, vehicles, and
human bodies are all well represented in generative training corpora.
We consider the harder case in which the target object has no
meaningful generative prior, and preserve it explicitly through
mask-locked compositing against the visible-surface mask of the
render.
The two designs trade off differently: unconstrained generation can
corrupt labels by adding or altering
instances~\cite{Karoly2025Mushroom}, whereas locked compositing keeps
labels exact and concentrates any residual appearance gap on the
object itself, whose realism is then governed by the fidelity of its
rendered texture, a trade-off we examine in
Section~\ref{sec:experiments}.

\paragraph{Gaussian splatting and VLM-driven scene construction.}
An alternative to procedural rendering is to reconstruct a scene from
real captures using 3D Gaussian Splatting.
Cut-and-Splat~\cite{Michiels2025CutSplat} trains a Gaussian-splatting
model of a target object from a video, automatically extracts the
object, and re-renders it onto random backgrounds with a
depth-guided pose, outperforming cut-and-paste and diffusion
baselines on their instance-segmentation benchmark.
Zanjani~\etal~\cite{Zanjani2025GSDataGen} use splatting to insert 3D objects
into reconstructed driving scenes for 3D detection.
These methods are visually strong but require dense captures of the
scene, which are rarely available in obscure detection domains.
On the scene-specification side,
SceneCraft~\cite{Hu2024SceneCraft} and
SEIG~\cite{He2026SEIG} show that LLM and VLM agents can synthesize
executable Blender code from a text description or a single image
respectively, via scene-graph planning with iterative library
learning or staged inverse graphics.
We use a similar VLM-to-Blender step as the entry point to our
pipeline, and defer photorealism to the subsequent LoRA-conditioned
diffusion pass rather than to the renderer.

\paragraph{Landmine detection and data scarcity.}
Surface-laid anti-personnel landmines pose acute detection challenges due to
their small size, camouflaging coloration, and high variability across
terrain and vegetation conditions~\cite{MineInsight2025}, motivating
robotic multi-sensor platforms for their detection and
disposal~\cite{Hasselmann2024MultiRobot}.
The MineInsight dataset~\cite{MineInsight2025} covers three off-road
vegetated tracks containing 15 inert landmines and 20 distractor
items, with synchronized monochrome, RGB, VIS-SWIR, LWIR (thermal),
and LiDAR streams from a UGV-mounted robotic arm, while
SULAND~\cite{Vivoli2024SULAND} offers ground-level RGB images of
PFM-1 (butterfly) and PMA-2 (starfish) landmines across Italian and
US sites.
Together, these benchmarks provide on the order of a few tens of
thousands of annotated frames and cover only a handful of terrain
types, far below the diversity needed for reliable generalization.
Vegetation height and density further degrade detection accuracy, as
quantified by Baur~\etal~\cite{Baur2024VegCoverage} for unmanned
aerial vehicle (UAV) sensing in minefield environments.
Collecting additional real data requires controlled environments with
inert ordnance and extensive safety protocols, making large-scale
annotation prohibitively expensive; even thermal detection of the
PFM-1 in vegetation operates under severely limited training
data~\cite{Malizia2025Thermal}.
Synthetic generation for this task has so far been limited to 2D
cut-and-paste composition of landmine templates onto real
backgrounds~\cite{Roboflow2025SynthPFM,AIBox2023PFM}, which can
neither reproduce the photometric coupling between a landmine and its
surroundings (cast shadows, ambient occlusion, inter-reflection) nor
embed the target geometrically \emph{within} the vegetation, precisely
the property identified above as decisive for
detection~\cite{Baur2024VegCoverage}.

\paragraph{Copy-paste and compositing augmentation.}
Copy-paste augmentation~\cite{Ghiasi2021CopyPaste} randomly composites
object instances onto new backgrounds and has become a standard
augmentation in instance segmentation.
Extensions apply Poisson blending~\cite{Perez2003Poisson} or
diffusion-based harmonization~\cite{Song2023ObjectStitch,Yang2023PaintByExample}
to reduce boundary artifacts.
Cut-and-paste strategies have been adopted for illegal landfill
detection~\cite{Lain2025Landfill} and other environmental monitoring
tasks where object appearance is stable but background diversity is
limited.
Our mask-locked compositing shares the motivation of preserving the
original object appearance, but operates in 3D: the occluding vegetation
is geometrically placed, the mask is exact, and the label derives from
the scene graph rather than from a 2D segmentation.

\section{Method}
\label{sec:method}

\begin{figure}[t]
\centering
\includegraphics[width=\linewidth]{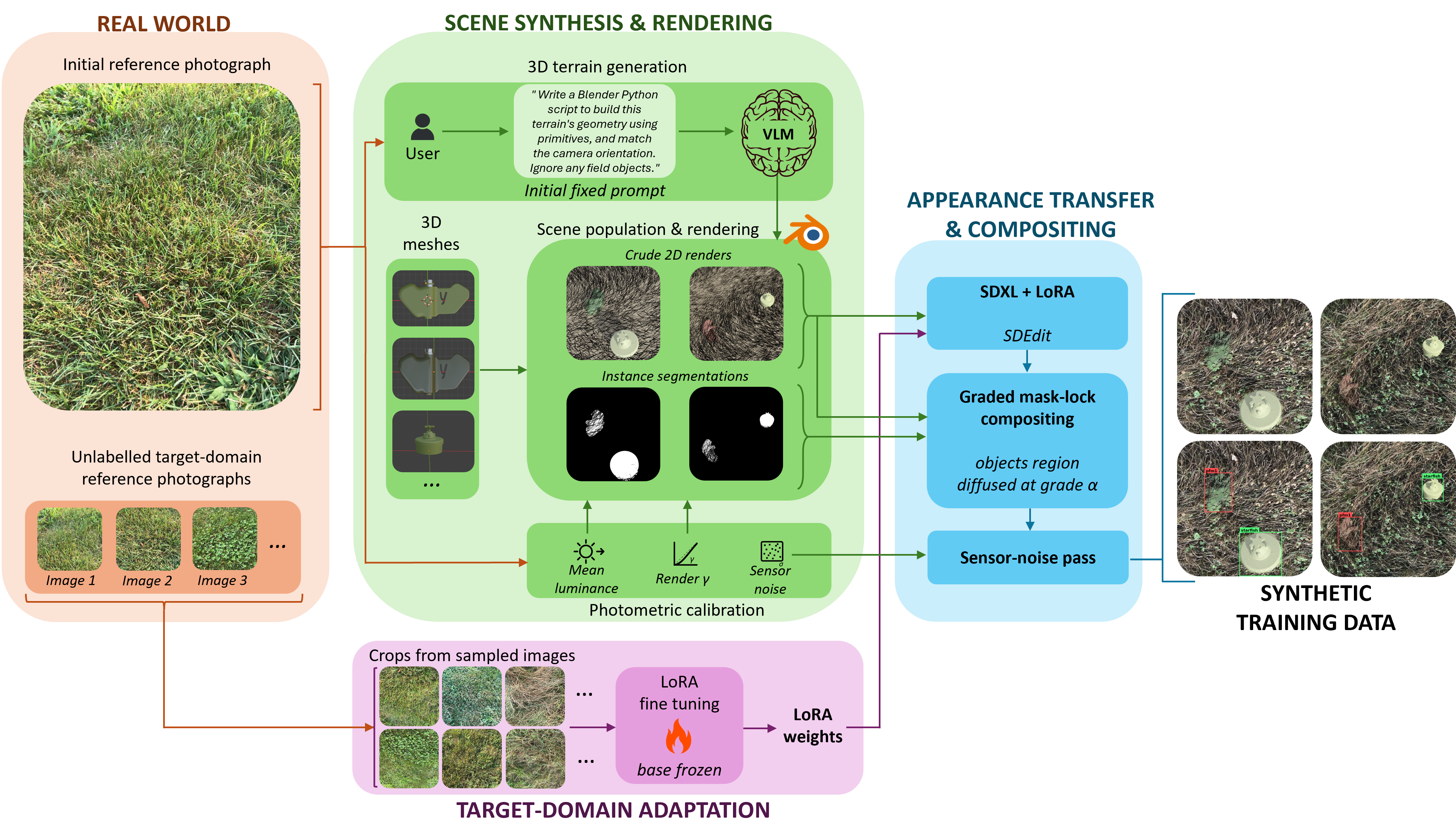}
\caption{Overview of the proposed pipeline.
The real-data budget is a handful of unlabeled photographs of the
target site.
\emph{One} of them serves as the scene reference: a vision--language
model emits a Blender Python script that approximates the terrain
geometry from primitives (\textsc{scene synthesis \& rendering}).
3D meshes of the target ordnance are placed in the resulting scene, so
bounding boxes, instance masks, and per-instance visibility are obtained
directly from the scene graph without manual annotation.
The \emph{full} photograph set feeds a one-time adaptation pre-phase,
prepared once per site and reused for every generated image: sun
strength, render gamma, and sensor noise are calibrated on the
photographs, and a LoRA fine-tuned on crops sampled from them adapts
the frozen diffusion prior to the target domain
(\textsc{target-domain adaptation}).
Each crude render is then re-textured by an SDEdit pass under the
LoRA-adapted prior, while a second copy bypasses diffusion and rejoins at
graded mask-lock compositing, where the object region is blended at
grade $\alpha \in [0,1]$; a final capture-realism pass matches the
target camera response.
Labels therefore remain valid by construction.
Boxes in the output denote the two target classes.}
\label{fig:pipeline}
\end{figure}

Figure~\ref{fig:pipeline} summarizes the pipeline; we describe each
stage below.

\paragraph{Stage 1: VLM scene primitives.}
We use a commercial vision--language model as the scene generator.
Prompted with one reference photograph, the VLM directly emits
Python code that instantiates the terrain in Blender as simple
primitives: ground palette, vegetation cards, and clutter (rocks,
sticks, patches of bare soil).
No pre-owned asset library is involved; at this stage the scene
consists of minimal primitives rather than a full reconstruction, and the
camera is subsequently sampled near the initial camera pose
estimated by the VLM.
3D meshes of the target ordnance are then placed by a scripted
spawner: several variants of the PFM-1 butterfly landmine and the PMA-2
anti-personnel landmine, sampled under a balanced class draw with
randomized yaw, position, and partial burial.
The spawner is fully parametric: target positions, class
distribution, spatial clustering, and burial can all be altered at
will, so the same scene supports arbitrary target layouts.
Scene lighting is likewise not hand-set: per-scene sun strength and
exposure are derived from the mean luminance of the reference
photographs; richer parameterizations (time of day,
sun position) are left to future work.
We deliberately do not ablate the VLM choice.
Commercial VLMs evolve on a shorter cadence than a research
pipeline, so a fixed-version claim would age within months; we
treat the VLM as a swappable scene-generation backend and report the
prompt template in Appendix~\ref{sec:appendix}.

\paragraph{Stage 2: adaptation pre-phase, unlabeled crops and terrain LoRA.}
This stage runs once per site, before any image generation; the
later stages only apply its output.
The pipeline's entire real-data budget is a handful of unlabeled
photographs, sampled from the existing distribution of the SULAND
dataset in the spirit of a generic robot mission over a field; the
scene reference of Stage~1 is one of them.
This rests on the working hypothesis that the field's appearance
stays approximately stationary across the site, with near-constant
vegetation; visually heterogeneous sites would require reference
photographs per appearance mode.
We take random square crops of each photograph, without bounding
boxes, masks, or class labels of any kind.
On these crops we fine-tune a LoRA adapter~\cite{Hu2022LoRA} on a
frozen Stable Diffusion XL backbone~\cite{Podell2023SDXL}.
The caption is a fixed short description of the target terrain and
is reused verbatim at inference.
The adapter is intentionally small so that it captures the
terrain's surface statistics rather than individual images; its
configuration is fixed across all experiments and reported in
Appendix~\ref{sec:appendix}.

\paragraph{Stage 3: mask-locked appearance transfer.}
We render one balanced pool of $1{,}000$ scenes and reuse it for
every ablation: the targets spawned in the 3D scene, their poses,
and their occlusions remain identical across all cells, so scene
variance is not a confound.
Passing through a 3D engine is what provides generic coverage of
the object in 3D and, importantly, full control over occlusion of
the target by vegetation at the initial stage, a property we
consider central to applications such as landmine detection and
target identification in vegetation.
Each render is passed through
SDEdit~\cite{Meng2022SDEdit}, prompted with the LoRA caption.
SDEdit re-noises the render into the diffusion latent space and
denoises it under the LoRA-conditioned prior, transferring surface
appearance while the 3D layout constrains structure.
For labels to remain valid, we then paste the rendered object
pixels back through the visible-surface mask exported by Blender.
We control how much diffusion may touch the object itself through a
graded mask-lock parameter $\alpha \in [0, 1]$: at $\alpha{=}0$ the
object is pixel-protected, at $\alpha{=}1$ the diffusion output is
kept everywhere.
Intermediate values feather the mask so that a thin, controllable
share of the object's appearance is inherited from the diffusion pass
while its geometry and label remain fixed.
In our experiments this grade emerges as an influential curation
choice (Section~\ref{sec:experiments}).
Figure~\ref{fig:lock-grid} shows representative outputs
across~$\alpha$.

\paragraph{Stage 4: sensor-noise pass.}
The composited image is finally passed through a light camera
model: we estimate the sensor-noise level of the reference
photographs~\cite{Immerkaer1996Noise} and inject noise so that the
generated images approximate the measured statistic; a fixed
compression pass completes the model.

\paragraph{Stage 5: detector training.}
We train an off-the-shelf YOLOv11-L
detector~\cite{Ultralytics2024YOLO11}, initialized from its
COCO-pretrained checkpoint~\cite{Lin2014COCO}, on the generated
dataset.
The training recipe follows the SULAND
benchmark~\cite{Vivoli2024SULAND} one-to-one, so the
synthetic-versus-real comparison in Section~\ref{sec:experiments} isolates
the training data as the only variable.
One consequence of rendering from a single scene deserves note:
because all generated images share the same background and asset
combinations, any validation split of them stays in-distribution,
and in-domain validation cannot indicate when
a model is ready to generalize
(Sec.~\ref{sec:experiments}).

The complete generation and training parameters, including the
SULAND augmentation policy, are tabulated in
Appendix~\ref{sec:appendix} together
with the VLM prompt template.

\section{Experiments}
\label{sec:experiments}

The experiments address three points:
(i)~whether the pipeline generalizes to real images, and how it
compares against a baseline trained on real labeled data under the
same training procedure (Sec.~\ref{ssec:transfer});
(ii)~which curation choice governs generalization: how much data
is used, or how the object itself is processed
(Sec.~\ref{ssec:ablation});
(iii)~whether anything cheaper than downstream evaluation predicts
generalization (Sec.~\ref{ssec:predictors}).

\subsection{Protocol}
\label{ssec:protocol}

The evaluation domain is the SULAND
benchmark~\cite{Vivoli2024SULAND}, which contains ground-level RGB
images of PFM-1 and PMA-2 surface-laid landmines across the ITA
and USA sites already introduced in Section~\ref{sec:related}.
The real labeled baseline (\textbf{R1}) is a YOLOv11-L
trained on the ITA split of SULAND, following the benchmark's own
recipe (details in Appendix~\ref{sec:appendix}); it represents the conventional
route in which labels collected at one site are reused for the
next.
All models, real and synthetic, are evaluated \emph{once} on the
USA split of SULAND ($4{,}436$~frames), which no model sees
during training.
This is a cross-site generalization setting: the terrain
distribution, capture rig, and lighting all shift between train
and test.
Metrics are macro-F1 (\textbf{mF1}) over the two landmine classes,
per-class recall, and mean average precision at
intersection-over-union (IoU) $0.5$ (\textbf{mAP50}).
Every synthetic configuration is trained on $1{,}000$ generated
images with class balance $790{:}790$ instances by design.
We report mean~$\pm$~standard deviation over five training seeds
per configuration.

\subsection{Generalization to real images}
\label{ssec:transfer}

Table~\ref{tab:main} lists the two baselines and the four levels of
the mask-lock grade $\alpha$.
The reference synthetic configuration (\textbf{M2}, $\alpha{=}0.10$)
reaches
$\text{mF1} = 0.584 \pm 0.025$,
against $0.393 \pm 0.021$ for the real
labeled baseline (\textbf{R1}).
The difference is several times the seed variability
($\approx 5.8$ combined seed standard deviations).
We read this as preliminary evidence, with the caveats of
Section~\ref{sec:conclusion}: on this benchmark and under this
recipe, the detector trained on our synthetic data generalized
better across sites than the one trained on the larger real
labeled set.
R1's absolute level also shows that the ITA-to-USA shift is
severe, which contextualizes how much headroom the label-reuse
route leaves on this benchmark.
The raw frame counts overstate the supervision gap: of the
$22{,}756$ real training frames, $5{,}234$ contain landmine
instances, against roughly $900$ of the $1{,}000$ synthetic images,
an effective ratio closer to $6{:}1$ than to $23{:}1$.
One access asymmetry deserves note: R1 sees no data from
the test site, whereas our pipeline consumes a handful of unlabeled
test-site photographs (the exact budget is swept in
Table~\ref{tab:budget}), so the comparison is training on
source-site labels alone versus unsupervised adaptation to the
target site.
Raw computer-graphics (CG) renders without the diffusion stage
(\textbf{R2}) reach only $0.194 \pm 0.111$: they teach the
renderer, not the world, and their seed-to-seed variance is an
order of magnitude larger than at any $\alpha > 0$.

\begin{table}[t]
\centering
\caption{Cross-site generalization on SULAND
(ITA / synthetic~$\rightarrow$~USA split; five seeds).
\textbf{R1} is the real labeled baseline, \textbf{R2} the raw
computer-graphics (CG) pool, and \textbf{M0}--\textbf{M3} sweep the
mask-lock grade~$\alpha$ over synthetic pools of the same size.
\textbf{Bold}: best value per column among the synthetic
configurations; values within a tie band ($\Delta < 0.01$ on
per-class recall, for which we do not report seed variance) are
bolded together and should not be read as ordered.}
\label{tab:main}
\small
\setlength{\tabcolsep}{4pt}
\begin{tabular}{@{}ll c c c c@{}}
\toprule
& Config & mF1 & R\,PFM-1 & R\,PMA-2 & mAP50 \\
\midrule
R1 & Real labeled (22.7k)                    & $0.393 \pm 0.021$          & $0.377$          & $0.217$          & $0.327 \pm 0.015$ \\
R2 & Raw CG, no diffusion                    & $0.194 \pm 0.111$          & $0.208$          & $0.185$          & $0.136 \pm 0.095$ \\
\midrule
M0 & Hard mask-lock ($\alpha{=}0$)           & $0.441 \pm 0.051$          & $\mathbf{0.579}$ & $0.196$          & $0.346 \pm 0.037$ \\
M1 & Graded, $\alpha{=}0.05$                 & $0.503 \pm 0.060$          & $0.571$          & $0.285$          & $0.401 \pm 0.049$ \\
M2 & Graded, $\alpha{=}0.10$ \emph{(ref.)}   & $\mathbf{0.584 \pm 0.025}$ & $\mathbf{0.577}$ & $\mathbf{0.418}$ & $\mathbf{0.481 \pm 0.024}$ \\
M3 & Unlocked ($\alpha{=}1$)                 & $0.466 \pm 0.030$          & $0.378$          & $0.339$          & $0.376 \pm 0.034$ \\
\bottomrule
\end{tabular}
\end{table}

\subsection{Which curation choice matters}
\label{ssec:ablation}

The mask-lock sweep in Table~\ref{tab:main} shows an inverted-U
across the four sampled values of~$\alpha$; a denser grid would be
needed to trace the curve between them.
Pixel-protecting the object~(M0) preserves labels but also
preserves the object's appearance gap: PMA-2 recall stays at the
real-data level ($0.20$) and mF1 does not clear the real baseline
within error.
Lightly diffusing the object~(M2, $\alpha{=}0.10$) transfers the
object's appearance too: PMA-2 recall roughly doubles from~$0.20$
to~$0.42$, while PFM-1 recall stays flat at $0.57$--$0.58$ across
M0--M2.
We speculate that the PFM-1's distinctive silhouette already
suffices for detection, whereas the low-contrast PMA-2 disc
depends on surface appearance, which is exactly what the graded
lock transfers.
Removing the lock entirely~(M3) dissolves parts of the object and
gives back most of the gain.
M3 also doubles as an unprotected-diffusion control, analogous in
spirit to full-regeneration pipelines such as K\'{a}roly and
Galambos~\cite{Karoly2025Mushroom}: the LoRA-adapted SDEdit pass
touches the whole frame, object included.
In our runs M3 falls short of M2, suggesting that a substantial
share of the gain comes from the graded lock rather than from the
adapted diffusion pass alone.
Among the axes we sweep, this one produces the largest shift.

\begin{figure}[!tbp]
\centering
\includegraphics[width=0.78\textwidth]{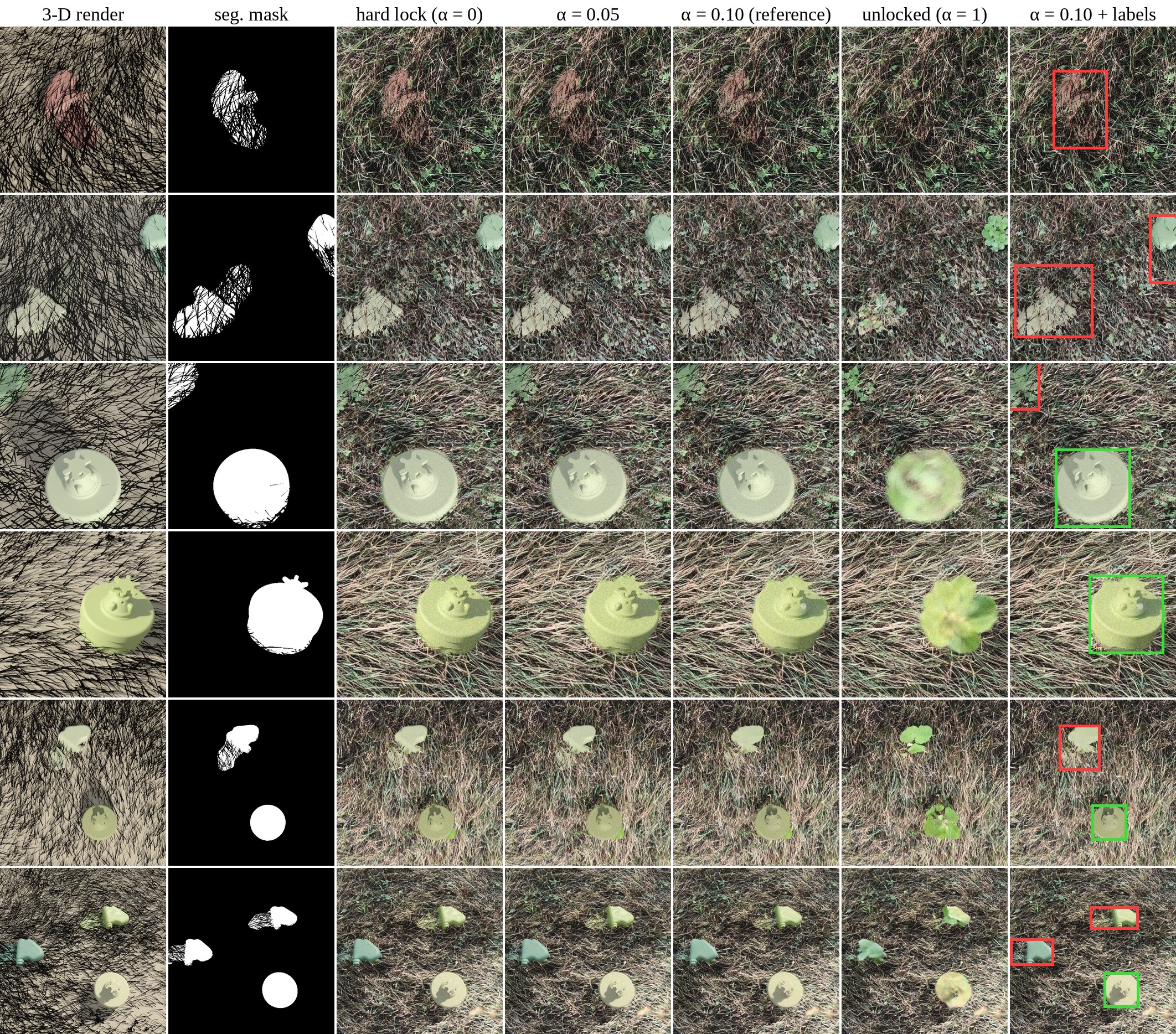}
\caption{Mask-lock grade $\alpha$ across six scenes (rows), same
diffusion seed per row.
Columns: 3D render, occlusion-aware segmentation mask, lock grades
in increasing order ($\alpha = 0$, $0.05$, $0.10$, $1$), and the
reference output with its 2D labels (red: PFM-1, green: PMA-2).
Rows 1--2: heavily occluded PFM-1s ($41\%$ and $51\%$ visible);
rows 3--4: PMA-2 close-ups; rows 5--6: multi-landmine scenes.
Unlocked, objects are absorbed into the background prior, most
completely where occlusion is heaviest.}
\label{fig:lock-grid}
\end{figure}

\begin{figure}[!tbp]
\centering
\includegraphics[width=0.78\textwidth]{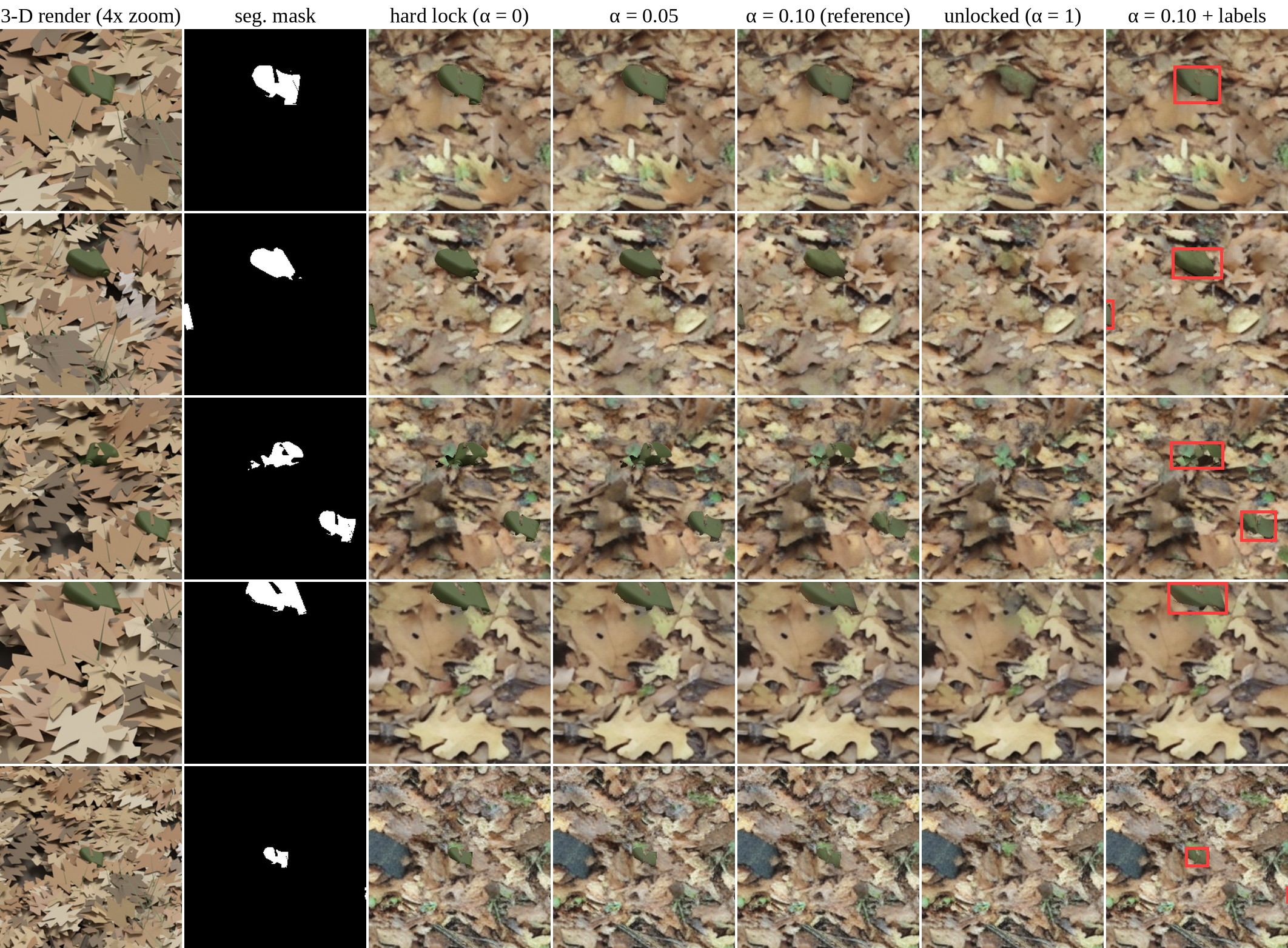}
\caption{The same comparison on the MineInsight forest-floor
terrain (RGB stream; five scenes; leaf-litter LoRA), same columns as
Figure~\ref{fig:lock-grid}; row~4 is a focal close-up at unchanged
camera position.
As on grassland, unconstrained diffusion absorbs the small object
into the background prior; grades were evaluated quantitatively on
SULAND only.}
\label{fig:lock-grid-mnst}
\end{figure}

\paragraph{Qualitative behavior of the lock.}
Figures~\ref{fig:lock-grid} and~\ref{fig:lock-grid-mnst} show
scenes from the shared render pool across the mask-lock
grade~$\alpha$, together with the scene-derived segmentation masks
and 2D labels.
Because the per-image diffusion seed is fixed, every column shows
the \emph{same} scene and the same diffusion draw; the only change
is how much of the object the diffusion may touch.
At $\alpha{=}0$ the object keeps its raw rendered appearance; at
$\alpha{=}0.05$--$0.10$ the object inherits surface statistics from
the terrain prior while its geometry and label are preserved; fully
unlocked ($\alpha{=}1$), the diffusion dissolves the object into
vegetation-like shapes and the label no longer marks a landmine.
Figure~\ref{fig:lock-grid-mnst} repeats the comparison on the
MineInsight forest-floor terrain~\cite{MineInsight2025}, using its
RGB stream only, with the same methodology and the same grades, as
a qualitative check that the behavior is not specific to grassland.

Table~\ref{tab:budget} sweeps three budget axes: reference
photographs ($1$, $5$, $10$), patches per photograph ($2$, $5$,
$14$), and generated image count ($100$, $250$, $500$, $1{,}000$).
The full sweep was run at the hard-lock setting ($\alpha{=}0$).
The smallest-budget level of each axis (one photograph, two patches,
one hundred images) was then re-measured at the reference lock
($\alpha{=}0.10$), five seeds each, as the most demanding test of
the flatness claim; the remaining combinations were not re-run for
compute reasons, as the second column of Table~\ref{tab:budget}
shows.
Within the measured cells, the photograph and patch axes are flat
within error; notably, a \emph{single} reference photograph at the
reference lock reaches $0.562 \pm 0.011$ mF1, within one standard
deviation of the ten-photograph configuration, and is the most
seed-stable cell in the study.
This presumes an approximately uniform terrain appearance across
the site, as holds for the grassland evaluated here; we return to
this assumption in Section~\ref{sec:limitations}.
The image-count axis is the exception: at the reference lock the
curve rises by ${\sim}0.07$~mF1 from $100$ to $1{,}000$ images
before flattening, still secondary to the $+0.14$ of the mask-lock
axis itself.
Within the ranges we could afford, quantity appears secondary to
how the object is processed.
A calibrated sensor-noise pass, ablated separately, leaves mF1
unchanged within seed variance at both lock settings (at
$\alpha{=}0$: $0.441$ vs.\ $0.448$; at $\alpha{=}0.10$:
$0.584 \pm 0.025$ vs.\ $0.571 \pm 0.048$), though
we keep it for photometric fidelity of the generated images.

\begin{table}[t]
\centering
\caption{Budget axes, five seeds per cell.
Full sweep at hard lock ($\alpha{=}0$); axis extremes re-measured
at the reference lock ($\alpha{=}0.10$); \emph{n/a} = cell not
re-measured.
The reference configuration (10~photographs, 14~patches, $1{,}000$
images) scores $0.441 \pm 0.051$ at $\alpha{=}0$ and
$0.584 \pm 0.025$ at $\alpha{=}0.10$ (Table~\ref{tab:main}).
\textbf{Bold}: best value per column among the rows shown.}
\label{tab:budget}
\small
\setlength{\tabcolsep}{8pt}
\begin{tabular}{@{}ll c c@{}}
\toprule
& Axis, level & mF1 at $\alpha{=}0$ & mF1 at $\alpha{=}0.10$ \\
\midrule
P1 & 1 photograph ($14$ crops)  & $0.485 \pm 0.070$          & $\mathbf{0.562 \pm 0.011}$ \\
P2 & 5 photographs              & $0.487 \pm 0.028$          & n/a               \\
\midrule
C1 & 2 patches / photograph     & $0.461 \pm 0.042$          & $0.564 \pm 0.023$ \\
C2 & 5 patches / photograph     & $\mathbf{0.505 \pm 0.042}$ & n/a               \\
\midrule
S1 & $100$ generated images     & $0.396 \pm 0.085$          & $0.514 \pm 0.042$ \\
S2 & $250$ generated images     & $0.410 \pm 0.084$          & n/a               \\
S3 & $500$ generated images     & $0.497 \pm 0.051$          & n/a               \\
\bottomrule
\end{tabular}
\end{table}

\subsection{Predictors of generalization}
\label{ssec:predictors}
Every synthetic configuration in Tables~\ref{tab:main}
and~\ref{tab:budget} reached $\text{mAP50} \geq 0.99$ on a
validation split of the generated images that the detector never
trains on (except the unlocked M3, which plateaued
at $0.84$--$0.87$, and one $100$-image seed at $0.91$), yet
real-domain mF1 spanned $0.05$ to $0.62$ across configurations and
seeds; as anticipated in Section~\ref{sec:method}, in-domain
validation was uninformative about generalization in our runs.
The spread is systematic rather than random: moving any
configuration from the hard lock to the reference lock roughly
halved its seed variance (\eg, $0.070 \rightarrow 0.011$ for the
one-photograph cell), so the graded lock appears to stabilize
generalization as well as improve its mean.
Within this study, curation decisions had to be guided by
downstream evaluation.

\section{Limitations}
\label{sec:limitations}

Five limitations bound the present findings.
First, the scene originates from a single reference image, and the
VLM's first-pass reconstruction might be imprecise; an agent-in-the-loop
refinement could reduce this imprecision.
Second, the camera distance to the ground is only approximately
recovered, while object scale matters for the graded lock.
Third, the terrain is assumed approximately uniform, as in the
grassland studied here and in the leaf-litter environment of
Figure~\ref{fig:lock-grid-mnst}; real sites may mix vegetation with
asphalt or bare
regions, which would call for a semantic-region analysis and,
eventually, a region-conditioned diffusion pass.
Fourth, the study covers one target family, one evaluation split,
and one detector, and deliberately does not ablate the 3D assets
or their fidelity.
Fifth, the mask-lock grade is sampled at four values only, so the
inverted-U of Section~\ref{ssec:ablation} should be read as a trend
rather than a fitted curve.

\section{Conclusion and future work}
\label{sec:conclusion}

We presented a preliminary study of a VLM-guided render-and-refine
pipeline for detection in vegetated terrain when labeled real data
are scarce, expensive, or dangerous to collect.
The pipeline turns a handful of unlabeled photographs of the site
and a small library of 3D meshes into a labeled detection dataset
with free bounding boxes, masks, and per-instance occlusion.
A graded mask-lock parameter interpolates between hard label
protection and unconstrained diffusion of the object; in our
experiments its middle regime generalized best.
At that setting, a detector trained on our synthetic data matched
or exceeded the same detector trained on a larger labeled dataset
of real images for cross-site generalization on the SULAND
benchmark~\cite{Vivoli2024SULAND}, with comparable seed variance.
If these findings hold beyond this benchmark, the per-site data
question for detection pipelines of this kind would shift from
collecting and labeling new frames to photographing the site and
regenerating the training set.

\paragraph{Scope of the claim.}
These are preliminary findings on one target class family, one
terrain type, one evaluation split, and one detector architecture.
We also deliberately do not ablate the use of 3D meshes or the
fidelity of the meshes themselves in the render.
The synthetic-versus-real comparison should therefore be read as
evidence that a small, curated synthetic set can be
competitive for cross-site generalization under this recipe, not
as a general claim about synthetic data superiority.

\paragraph{Future work.}
Four directions follow from the limitations we observed.
\emph{Richer 3D curation}: the VLM currently emits only minimal
scene primitives, and substantial work remains on capturing the
initial 3D distribution of a site inside the simulator; supporting
the capture with range sensing (\eg, a LiDAR back-support) could
standardize the scene geometry and remove the dependence on any
particular VLM, which we cannot meaningfully ablate here.
\emph{Better blending}: the mask-lock stage today paints the
rendered object back through a binary visible-surface mask;
soft harmonization and learned compositing could remove the small
residual seam that remains at low~$\alpha$ and may absorb the
appearance gap the graded lock currently transfers.
\emph{Higher-fidelity meshes}: the object realism is upper-bounded by
the mesh textures, especially for weathering, dirt, and paint wear
that are absent from our meshes; capturing physically-based
materials from inert ordnance would move the object closer to the
target distribution without further diffusion pressure.
\emph{Broader evaluation}: a second detector family, additional
terrain pairs (forest, sand, mixed), a denser mask-lock grid, and
a cheap and calibrated predictor of generalization are natural
next steps.

\ifreview
\section*{Acknowledgments}
Omitted for blind review.
\else
\section*{Acknowledgments}
\FundingAcknowledgement
\\The authors used Claude Fable 5 and GPT-5.6 Sol (ChatGPT) solely to polish the initial human-written manuscript for improved readability, clarity, and flow; all scientific content and conclusions remain the responsibility of the authors.
\fi

\bibliographystyle{splncs04}
\bibliography{main}

\appendix

\section{Appendix}
\label{sec:appendix}

\subsection{Configuration}
\label{ssec:app-config}

\paragraph{Generation.}
Blender (Cycles) renders at $960{\times}960$~px, 48~samples;
1--3~landmines per scene with a stratified $1{:}1$ class draw
(4~PFM-1 mesh variants, 1~PMA-2), $10\%$ landmine-free negative frames;
camera poses are rejected unless every frustum corner lands on
vegetated ground.
The draw balances classes over the whole pool: $900$ of the
$1{,}000$ frames carry $1{,}580$ landmine instances in total,
$790$ per class, and the remaining $100$ frames are negatives.
Per-instance visibility is computed from paired visible/full
silhouette renders (pool median $0.96$; $11\%$ of instances below
$0.70$).
The terrain LoRA uses rank~$12$, $1{,}000$~steps, AdamW at
$10^{-4}$, $512$-px crops, and the fixed caption
\emph{``sulandterrain field ground terrain, green and dry grass,
clover, top-down photograph''}, reused verbatim at generation time.
SDEdit runs at strength~$0.42$, $35$~steps, classifier-free
guidance~$6.0$, LoRA scale~$0.8$; the per-image noise generator is
seeded with $42+\mathrm{image\ id}$, which makes every ablation cell
pixel-aligned to every other.

\paragraph{VLM scene generation.}
The scene-generation backend is Claude Fable~5 (Anthropic; model id
\texttt{claude-fable-5}).
A single prompt provides the reference photograph and asks the
model to emit Python code that instantiates the terrain in Blender
as scene primitives (ground, vegetation cards, clutter); no
pre-owned asset library is involved.
The initial fixed prompt, also shown in
Figure~\ref{fig:pipeline}, is: \emph{``Write a Blender Python
script to build this terrain's geometry using primitives, and
match the camera orientation. Ignore any field objects.''}

\paragraph{Capture realism.}
Per-scene sun strength follows the reference photograph's mean
luminance ($e_0 \cdot 2.5\,(0.6 + 1.4\bar L)$), and each render is
gamma-matched to its own reference photograph's mean luminance
(median $\gamma = 0.42$).
The sensor-noise target measured on the reference photographs is
$\sigma = 12.3$ (0--255 scale); closed-loop calibration of the
injected noise against the final JPEG (quality~88) converged at an
injection of $\sigma = 15.8$, leaving a final synthetic statistic of
$12.6$.

\paragraph{Detector training.}
Table~\ref{tab:params} lists the complete training configuration:
the SULAND benchmark recipe, used unchanged for the real baseline
and for every synthetic configuration.
All rendering, LoRA fine-tuning, diffusion, and detector training
ran on a single NVIDIA RTX~5090 (32~GB).

\begin{table}[t]
\centering
\caption{Complete detector-training configuration (SULAND recipe,
identical for all runs) and generation constants.}
\label{tab:params}
\small
\begin{tabular}{@{}llll@{}}
\toprule
\multicolumn{2}{@{}l}{Training} & \multicolumn{2}{l@{}}{Augmentation (SULAND)} \\
\midrule
model & YOLOv11-L & mosaic & 1.0 \\
epochs & 100 & hsv (h/s/v) & 0.015 / 0.7 / 0.4 \\
patience & 25 & translate & 0.1 \\
input size & 640 & scale & 0.5 \\
batch & 32 & horizontal flip & 0.5 \\
optimizer & AdamW & mixup / copy-paste & 0 / 0 \\
initial lr & $3{\times}10^{-4}$ (cosine) & degrees / shear & 0 / 0 \\
weight decay & $5{\times}10^{-4}$ & \multicolumn{2}{l@{}}{deterministic mode on} \\
dropout & 0.1 & \multicolumn{2}{l@{}}{seeds \{42, 1, 2, 3, 4\}} \\
\midrule
\multicolumn{2}{@{}l}{Generation} & & \\
\midrule
SDEdit strength & 0.42 & LoRA rank / steps & 12 / 1{,}000 \\
diffusion steps & 35 & LoRA lr / crops & $10^{-4}$ / 512~px \\
guidance & 6.0 & noise inject $\sigma$ & 15.8 (measured) \\
LoRA scale & 0.8 & negatives & 10\% \\
\bottomrule
\end{tabular}
\end{table}

\subsection{Quality-metric probes}
\label{ssec:app-probes}

We also probed two supporting signals, not used to guide the study
but included here as sanity checks: (i) image-level distributional
realism, measured as Kernel Inception Distance
(KID)~\cite{Binkowski2018KID} between each generated set and real
target-domain crops, gave a Spearman correlation of
$\rho = -0.19$ ($p = 0.60$) against downstream mF1 across
configurations; it isolates the raw-CG pool as an outlier but does
not rank the diffused sets.
(ii) An open-vocabulary detection probe
(Grounding-DINO~\cite{Liu2024GroundingDINO} recall on
the target class) reported near-saturated recall (0.88 to 0.98) for
every generated set, including the raw-CG pool that generalizes
worst ($\rho = -0.39$, $p = 0.27$).
Both are observations of this study's setting; we did not tune
either probe, and different prompts, thresholds, or backbones could
behave differently.

\subsection{SULAND protocol details}
\label{ssec:app-suland}

The ITA split provides $22{,}756$ labeled training frames; the USA
split used for evaluation provides $4{,}436$ frames.
Evaluation uses the standard Ultralytics validation routine at
IoU~$0.5$, run once per trained model; per-class F1 is read at the
F1-optimal operating point of the precision--recall curve,
computed identically for all models, and macro-F1 is the mean of
the two per-class F1 scores.

\subsection{Real-baseline provenance}
\label{ssec:app-r1}

The real baseline is trained on the SULAND ITA split:
$22{,}756$~frames, of which $5{,}234$ contain landmine instances
($5{,}510$~boxes, ${\approx}1{:}1$ between classes), by the same
training configuration as above.
The ten unlabeled adaptation photographs are drawn from the
USA-site (evaluation) distribution; each was matched to its source
evaluation frame (feature cosine $> 0.95$, one per capture
sequence).
Removing the matched frames from the evaluation set is a null
effect on the model the concern applies to, the synthetic
configuration whose LoRA saw the photographs: the reference M2
scores mF1 $0.584 \pm 0.025$ and mAP50 $0.481 \pm 0.024$ on the
full set, versus $0.584 \pm 0.025$ and $0.482 \pm 0.024$ with the
matched frames removed (five seeds).
Evaluation therefore uses the full set, and the overlap is
disclosed rather than excluded.
Frames from the same capture sequences share appearance with the
adaptation photographs by construction; that is the
site-adaptation setting itself rather than duplicate leakage,
which the exclusion above bounds.

\end{document}